\documentclass[runningheads]{llncs}
\usepackage[T1]{fontenc}
\usepackage{graphicx}
\usepackage[caption=false]{subfig}
\usepackage[export]{adjustbox}

\usepackage{siunitx}
\usepackage{booktabs}
\usepackage{tabularx}
\usepackage{makecell}
\usepackage{colortbl}
\usepackage{multirow}
\usepackage{tablefootnote} 

\usepackage{xcolor}
\usepackage{color}

\usepackage{hyperref}
\usepackage{enumitem}
\usepackage{listings}
\usepackage{float}
\usepackage{cite}
\usepackage{listings}
\usepackage{enumitem}
\usepackage{orcidlink}

\newcommand{\sent}[1]{``\emph{#1}''}

\newcolumntype{C}{>{\centering\arraybackslash}X}
\newcolumntype{R}{>{\raggedleft\arraybackslash}X}
\newcolumntype{L}{>{\raggedright\arraybackslash}X}
\newcolumntype{P}[1]{>{\raggedleft\arraybackslash}p{#1}}

\newcommand{\rowheader}[2]{\parbox[t]{8mm}{\multirow{#1}{*}{\rotatebox[origin=c]{90}{\shortstack{#2}}}}}

\newcommand{\mypar}[1]{\vspace{0.5pt}\noindent\textbf{#1.}}

\begin{document}
\title{Assisted Data Annotation for Business Process Information Extraction from Textual Documents}
\titlerunning{Assisted Business Process Information Annotation}
%
\author{%
    Julian Neuberger\inst{1}\orcidlink{0009-0008-4244-7659} 
    \and Han van der Aa\inst{2}\orcidlink{0000-0002-4200-4937}
    \and Lars Ackermann\inst{1}\orcidlink{0000-0002-6785-8998} 
    \and Daniel Buschek\inst{1}\orcidlink{0000-0002-0013-715X}
    \and Jannic Herrmann\inst{1}
    \and Stefan Jablonski\inst{1}
}
\authorrunning{J. Neuberger et al.}
%
\institute{%
    University of Bayreuth, Bayreuth, Germany\\
    \email{firstname.lastname@uni-bayreuth.de}
\and
    University of Vienna, Vienna, Austria\\
    \email{han.van.der.aa@univie.ac.at}
}
\maketitle              
\begin{abstract}
Machine-learning based generation of 
process models from natural language text process
descriptions provides a solution for the time-intensive 
and expensive process discovery phase. Many organizations
have to carry out this phase, before they can utilize 
business process management and its benefits. Yet, 
research towards this is severely restrained by an apparent 
lack of large and high-quality datasets. This lack 
of data can be attributed to, among other things, 
an absence of proper tool assistance for dataset 
creation, resulting in high workloads and inferior data 
quality. We explore two assistance features to support 
dataset creation, a recommendation system for identifying 
process information in the text and visualization of the 
current state of already identified process information as 
a graphical business process model. A controlled user study 
with 31 participants shows that assisting dataset creators 
with recommendations lowers all aspects of workload, up to 
$-51.0\%$, and significantly improves annotation quality, 
up to $+38.9\%$. We make all data and code available to 
encourage further research on additional novel assistance 
strategies.

\keywords{Business Process Management 
\and Process Information Extraction
\and Natural Language Processing 
\and Human Computer Interaction}
\end{abstract}

\section{Introduction}\label{sec:introduction}

Business process management (BPM) can provide organizations 
with many benefits by improving their regular operating
procedures. Organizations looking to utilize these benefits 
first need to discover and model their business processes,
which is a very time consuming, and therefore expensive task
\cite{friedrich2011process}. To alleviate this, researchers 
in the BPM community use the information contained in natural 
language process descriptions from sources like quality management 
handbooks, documentation of standard operating procedures, or 
employee notes to automatically generate formal process models. 
While this area is actively researched~\cite{friedrich2011process,
ferreira2017semi, van2019extracting,ackermann2021data,
bellan2022extracting,neuberger2023beyond}, new and innovative 
approaches are quite rare. One reason is the limited availability 
of data to develop, train, and assess approaches. 
Recent initiatives aim to mitigate this
issue, providing a gold-standard dataset for the process
information extraction task---PET~\cite{bellan2023pet}.
With this dataset, systems for extracting process information
can be developed, which in turn are the basis for subsequent
automated model generation methods, allowing fully automated
process model generation from process descriptions.
Still, this dataset contains only 45 process descriptions,
which is not enough to train deep neural networks
\cite{neuberger2023beyond}, although they have been shown to be well 
suited for similar tasks in other areas~\cite{ackermann2023bridging}. 
Even techniques based on pretrained large language models are 
affected by the lack of data, as rigorous evaluation on many
different data sources is essential to assess their practicality, 
especially in light of the large variation in terms of the structure, 
style, and contents of textual documents that contain process 
information~\cite{van2015fragmentation}.

\begin{figure}[hbt]
    \centering
    \includegraphics[width=\linewidth]{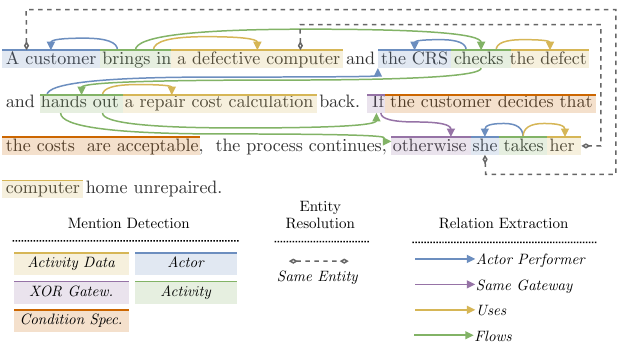}
    \caption{The first two sentences of document \textit{doc-1.2} in the 
    PET dataset, fully annotated with entity mentions, entity references,
    and relations.}
    \label{fig:motivating-example}
\end{figure}

The lack of suitable data for process information extraction tasks 
can in part be attributed to the effort required to establish gold 
standard annotations. Such annotations are a critical requirement 
for both training and evaluation of information extraction approaches.
However, manually annotating process information in textual 
process descriptions involves elaborate guidelines~\cite{bellan2023pet} 
and considerable ambiguity~\cite{van2018checking,franceschetti2023ambiguity}, 
making it time consuming and mentally taxing.
Fig.~\ref{fig:motivating-example} shows two sentences of a process 
description from the PET dataset, fully annotated with 
the gold-standard process information. Note, that annotating these sentences
requires identifying 14 process-relevant elements, and 16 dependencies between 
them, in just these two sentences, where the average description 
in PET has 9.27 sentences~\cite{bellan2023pet}. We discuss the task in detail in 
Sect.~\ref{sec:motivation} and how to circumvent this complexity in
Sect.~\ref{sec:workflow}. Additionally, depending on the annotation 
schema, some of these annotations are not intuitive, e.g., \sent{decides}, 
which would intuitively be annotated as an \textit{activity}, underlining 
the need for annotation guidelines mentioned above.

Recognizing this issue, we explore how dataset creators (\textit{annotators}) 
can be assisted in their data annotation task, so that their workload is
lightened, while simultaneously improving the quality of their extractions.
Therefore, in this paper, we propose and evaluate the benefits of two assistance 
features that can support human annotators: (1) AI-based recommendations, which 
allow annotators to quickly tackle trivial parts of the annotation task---as well 
as receive suggestions for less trivial aspects---and (2) the use of a visualizations 
of the currently annotated information through a graphical process model, which 
allows annotators to observe the process that they have so far captured.
Note that, although the task of text annotation is generic, the assistance 
features are tailored to the specifics of text annotations for process 
information extraction.

We implement both assistance features in a prototypical annotation tool 
that we use as a basis for a rigorous user study with 31 participants, 
ranging from modeling novices to experts, to assess the effectiveness and 
efficiency of the proposed features. Code%
\footnote{see~\url{https://github.com/JulianNeuberger/assisted-process-annotation}} 
and data\footnote{see~\url{https://zenodo.org/doi/10.5281/zenodo.12770686}} are made 
publicly available 
to the research community, to allow others to efficiently and effectively 
annotate their own datasets, but also encourages exploration of additional
assistance features. The main insights from this study are as follows.

\begin{enumerate}[topsep=0pt]
    \item Assisting annotators with suggestions made by artificial intelligence systems is observed to make annotating process relevant information in 
textual process descriptions significantly easier. This results in a significant reduction of key workload metrics by more than one half ($-51.0\%$). At the same time, assistance improves the quality of 
extracted information measured in $F_1$ score by up to $+0.224$ ($+38.9\%$). 

\item The use of assistance features is recognized to considerably reduce the gap between novice and experienced process modelers in annotation tasks. Specifically, complete beginners can reach the comparable annotation quality as expert annotators, speeding up the training process for new data annotators considerably. This insight shows that annotation features can help  assembling larger data annotation teams, and speeds
up the creation of new datasets in the space of business process 
model extraction from natural language text.
\end{enumerate}
\noindent The rest of this paper is structured as follows. In Sect.~\ref{sec:related-work}, we discuss related work on process information extraction, relevant user studies, and annotation tools.
In Sect.~\ref{sec:concept}, we present our concept behind a tool built
specially for annotating textual process descriptions, its implementation
in a research prototype, and the assistance features. In Sect.~\ref{sec:study-design}
we describe the design and execution of the user study. We present results 
for this study in Sect.~\ref{sec:results}.
We conclude the paper in Sect.~\ref{sec:conclusion}, summarizing, discussing 
limitations, and describing future work.

\section{Related Work}\label{sec:related-work}

Work related to this paper can be roughly categorized into three sections.

\mypar{Business process information extraction} 
The last decades have seen
various approaches to the task of extracting process relevant information
from natural language text, including systems based on expert-defined rules
\cite{ferreira2017semi,van2019extracting,bellan2022extracting,quishpi2020extracting},
data-driven ones~\cite{ackermann2021data,bellan2023pet,neuberger2023beyond}, 
and systems based on pretrained generative large language models
\cite{kourani2024process,bellan2022extracting}. We use approaches from 
\cite{neuberger2023beyond} and~\cite{bellan2023pet} in our work to implement
annotation recommendations. 
Many of the works mentioned also propose data annotation schemata tailored towards 
specific modeling languages, such as \cite{quishpi2020extracting,van2019extracting} 
for declarative process modeling, or towards different task descriptions, such as
\cite{Qian2020} for information extraction from process relevant sentence fragments.

We focus on PET, as it is heavily biased towards the current industry standard, BPMN,
and the to date largest available dataset.
PET was extended with the notion of entity identities~\cite{neuberger2023beyond}, 
i.e., the task of resolving multiple mentions of the same process element across 
the textual description to a single one. This is important for properly modeling 
business objects and process participants, which would otherwise be duplicated in 
the generated model. In this paper we use this extended version of PET.

\mypar{User studies}
Rosa et al.~\cite{rosa2022visual} develop and evaluate a tool for business 
process modelling
which assists users by identifying core BPMN 2.0\footnote{
\url{https://www.omg.org/bpmn/}, accessed July 4, 2024.} elements and
highlighting them in the process description. Our work, in 
contrast, aims to be a step towards alleviating the data scarcity problem
in business process model generation from text, by making data annotation
easier. In a study of similar size to ours, the authors of the \textit{BPMN Sketch Miner}
evaluate its usefulness to support process modelling based on textual descriptions
with visual representations of process elements~\cite{ivanchikj2020text}. 
While their study mainly focuses on usability and 
subjective values, ours also considers objective measures.

\mypar{Annotation tools}
Both our concept and implementation for assisted process-relevant information 
annotation are related to a number of annotation tools. 
These can usually be used to annotate text for use in
Named Entity Recognition (NER), Entity Matching and Resolution (ER), or 
Relation Extraction (RE), tasks which are similar to business process 
information extraction (compare definition in Sect.~\ref{sec:concept}).
Still, these tools are not designed with characteristics of business 
process descriptions in mind, including but not 
limited to, the high information density present in such descriptions, its
inherent ambiguity (see Sect.~\ref{sec:introduction}), and the target 
down-stream task, i.e., generation of a formal and graphical process model.
Additionally, unlike in the NLP community, data annotators for process 
information extraction are often times experts in BPM, but not in NLP, and
therefore can benefit from purposeful simplifications in the annotation tool.
In the following we will describe several notable examples of multi-purpose 
Natural Language Processing (NLP) data annotation tools, from which we drew
inspiration and how our proposed concept differs from them.

Doccano~\cite{doccano} provides features useful for collaborative data 
annotation, creating datasets in multiple languages, and comparing 
annotations between annotators. 
Label Studio~\cite{labelstudio} supports more 
machine learning domains, e.g., computer vision, and audio processing. This 
makes the tool even more multi-purpose and less bespoke, compared to our 
research prototype. The authors already have experience annotating textual 
business process descriptions using Label Studio~\cite{ackermann2021data},
which is integrated into our concept for assisted annotation (Sect.
\ref{sec:concept}).
Finally, INCEpTION~\cite{klie2018inception} uses
\emph{recommenders} to make suggestions for new annotations, which would 
fit our requirement for AI-based annotation recommendations, but to the 
best of the author's knowledge can not be extended to show the current state
of annotation as a BPMN model. The authors of~\cite{klie2018inception} did not 
investigate the effectiveness
of recommendations for text annotation, and while a positive effect seems
plausible, we are interested in proving and quantifying this effect.

\section{Concept for Assisted Annotation}\label{sec:concept}

This section outlines our concept for assisted data annotation. 
First, we define the task human annotators have to complete in 
Sect.~\ref{sec:motivation}. 
Based on this we motivate the need for more efficient and effective
data annotation and derive 
assistance features in Sect.~\ref{sec:assistance-features}.
Finally, we describe our research prototype implementation in
Sect.~\ref{sec:implementation}.

\subsection{The Process Information Extraction Task}\label{sec:motivation}

Ultimately, human annotators have to complete the process information
extraction task to annotate process descriptions with process-relevant
information. Therefore, we define this task in the following.
Consider, for example, document \textit{doc-1.2} from the 
PET dataset describing the process of a computer repair. 
Fig.~\ref{fig:motivating-example} shows this document fully annotated with
all process relevant information, which consists of three major categories. 
First, \emph{\textbf{Mentions}} of process relevant entities in PET are 
continuous sequences of text with a given type, for example, \emph{Actors} 
(process participants, ``a customer''), \emph{Activity Data} (business objects, 
``a computer''), or \emph{XOR Gateways} (decision points, often indicated by 
``if'', ``otherwise''). The last example illustrates, why we call detecting 
and extracting such mentions \emph{Entity Mention Detection} (MD) and not 
\emph{Named Entity Recognition} (NER). Named Entities are defined by either 
proper names (e.g., persons, locations) or natural kind terms (e.g., enzymes, 
species)\cite{li2020survey}. ``If'' or ``otherwise'' do not fall into this 
definition, which is why we use the more relaxed definition of (non-named) 
entities and the detection of their mentions within the text~\cite{xu2017local}.
Mentions are then resolved to \emph{\textbf{Entities}}, i.e., clustered, 
allowing subsequent model generation steps to only render a single process 
element, instead of multiple (one for each of its mentions). This task is 
called \emph{Entity Resolution} (ER) and is closely related with co-reference 
and anaphora resolution \cite{sukthanker2020anaphora}. \emph{\textbf{Relations}} 
between mentions define how these elements interact with each other. PET 
defines, for example, \emph{Flow} (order of task execution), \emph{Uses} 
(association between a task and the business object it uses), or 
\emph{Actor Performer} (assigning a process participant as executor of a 
given task).

\subsection{Annotation Workflow}\label{sec:workflow}

As we discussed in Sect.~\ref{sec:introduction}, annotating the process 
relevant information contained in textual process descriptions is a complex 
task and very demanding for the human performing the annotation, as it 
requires attention to three sub-tasks, as outlined in the previous section
\ref{sec:motivation}. We therefore split the task into its sub-tasks MD, ER, 
and RE. While this partially alleviates the issue of complexity, it will also 
allow us to assist annotators in these sub-tasks differently, and 
analyze how assistance features help during a specific sub-task. 
Fig.~\ref{fig:workflow-architecture} depicts the resulting workflow. After the 
annotator submits a natural language process description, they are then asked 
to select mentions (MD), resolve entities (ER), and define relations between
mentions (RE), in three separate steps. Finally, all information is shown
again, so that the annotator may reconcile any errors. 

While this workflow reduces the complexity of process information annotation 
by splitting it up into smaller tasks, the overall complexity remains high.
High density of information makes annotating very confusing, especially
for beginners. The example in Fig.~\ref{fig:motivating-example} contains a 
total of 40 words, of which only eight are not part of one of the 14 entity 
mentions (20\%), while also containing 14 relations between them. From previous
annotation experience in other tools (see Sect.~\ref{sec:related-work}), we 
know that this can be partially mitigated by splitting the task into smaller 
sub-tasks, e.g., focusing on a subset of entity and relation types, or by 
annotating the categories from above one after the other. Based on this
experience we defined a \emph{workflow}, which we describe in Sect.~\ref{sec:workflow}. 
High information density and the resulting complexity
of displaying this information also motivates us to find ways to visualize 
the information better, and help the user focus on information they potentially 
would miss otherwise. This results in two assistance features, 
\emph{visualization} and \emph{recommendation}, which we describe in 
Sect.~\ref{sec:assistance-features}.

\subsection{Assistance Features}\label{sec:assistance-features}

In one of our preliminary studies two assistance features were identified as
promising candidates for improving the efficiency, quality, and user experience of 
annotation documents for the process information extraction task. 

\mypar{AI-based annotation recommendations}
Building on the progress that has already been made in the development of automated information extraction approaches
for mentions, entities, and relations, 
we can present the user with recommendations for these elements. Interviews with BPM 
experts during the preliminary study and our review of related work (Sect.~\ref{sec:related-work})
suggested that recommendations can be a powerful tool 
for speeding up annotation in trivial cases and provide useful ideas in non-trivial ones.
We used an approach based on conditional random fields for extracting mentions, as 
presented by Bellan et al.~\cite{bellan2023pet}, with code from~\cite{neuberger2023beyond},
a pretrained neural co-reference resolver for entities~\cite{neuberger2023beyond}, and
a relation extraction approach based on gradient boosting on decision trees
\cite{neuberger2023beyond}. All approaches were trained with 80\% of the available data
(39 documents) and the rest was held out for use during the user study.
Recommendations are shown during the appropriate workflow steps and 
can be confirmed, discarded, edited, or marked for later review.

\mypar{Visual result representation}
Second, the information that a human annotator marks in a textual 
process description is always process relevant, i.e., a perfect
annotation results in a model that perfectly reflects the process
description. This shows how a human annotator may benefit from a
graphical process model as a visualization of the currently annotated 
information, as any missing information is reflected in the (therefore incomplete) 
graphical process model.
Visualizing the current state of annotation involves three major 
stages. First, in the \emph{\textbf{Consolidation}} stage, we assign conditions 
to their respective paths in the process, merge mentions of entities, and find 
the closest actor in the text left of activities that are not explicitly assigned
one. In the second stage, the \emph{\textbf{Vertex}} stage, we create process 
elements for all mentions, e.g., \emph{Tasks}, \emph{Data Objects}, \emph{Swimlanes}, 
etc. The final \emph{\textbf{Linking}} stage connects related elements, e.g., 
successive tasks and gateways with \emph{Sequence Flows}, if they are located in the same Pool, 
or \emph{Message Flows} between them. We also create \emph{Data Associations} between 
Data Objects and Tasks, adding the label of the Data Object to the label of the Task, 
for labels like ``\emph{send a mortgage offer}''. In this way the graphical process
model is generated and layouted automatically, and as such has limitations that
might affect its usefulness, which we discuss in Sect.~\ref{sec:limitations}.

\subsection{Implementation}\label{sec:implementation}

We have implemented our concept in a usable research prototype. It consists of 
a user-facing web application, 
implemented in JavaScript, using React\footnote{\url{https://react.dev/}, last 
accessed July 11, 2024.}. 

A backend server provides NLP pre-processing functionality, such as tokenization 
and the information extraction approaches for the recommendation assistance feature. 
It is implemented in Python 3.11, using spaCy\footnote{\url{https://spacy.io/}, 
last accessed July 11, 2024.} for pre-processing. When the user inputs a textual 
process description, it is first sent to this server to pre-process the text. The 
result is then displayed in the web application. In each of the annotation 
sub-tasks defined in Sect.~\ref{sec:workflow} the relevant information is 
extracted by the backend server and presented to the annotator as recommendations. 
The backend server is also responsible for storing annotation results. 
A second backend server is used for visualizing the current annotations by generating 
a formal process model in BPMN and its graphical representation. We implement this
using Java 17 and the Camunda Model API\footnote{
\url{https://docs.camunda.org/manual/7.21/user-guide/model-api/bpmn-model-api/},
last accessed July 11, 2024.}.

\begin{figure}[bt]
    \centering
    \includegraphics[width=\textwidth]{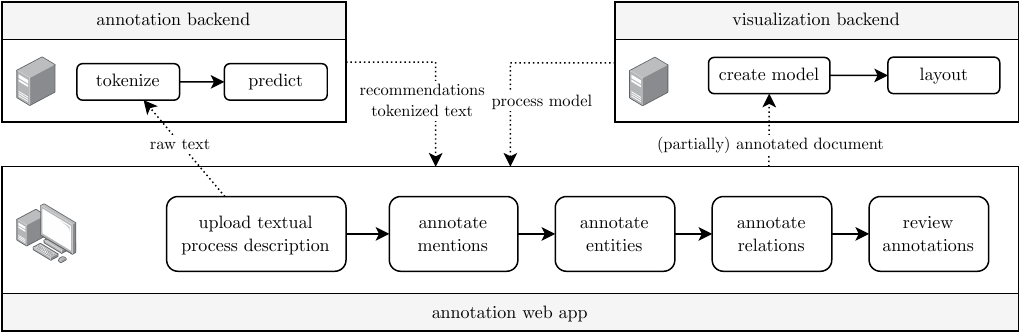}
    \caption{Visualization of the workflow and general architecture of our 
    implementation.}
    \label{fig:workflow-architecture}
\end{figure}

Both back-end servers expose any functionality using REST interfaces, which the 
user-facing web application can query. Fig.~\ref{fig:workflow-architecture} shows
the workflow in the web application, as well as the communication between servers.

\section{Study Design}\label{sec:study-design}

We conducted a user study with 31 participants to assess the effectiveness 
of the two assistance features, based on several measures in scenarios of 
varying assistance. Both measures and scenarios are defined later in this
section. We focus our efforts on answering the following three key research
questions.

\begin{enumerate}[topsep=0pt,itemindent=.7cm, label=\textbf{RQ\arabic*}]
    \item\label{rq:workload-effects} Which annotation assistance features
    or combinations lower the workload of annotating process information? 


    \item\label{rq:quality-effects} Which annotation assistance features
    or combinations improve the quality of annotations?

    \item\label{rq:bridge-gap} Which annotation assistance bridge the
    gap in annotation quality between beginner annotators and those with
    BPMN experience?
\end{enumerate}

\noindent The general setup of this study is as follows.
All supplementary material, such as the questionnaires
and resulting data can be found online, see Sect.~\ref{sec:introduction}.

\mypar{Study procedure}
Each participation within our user study, is structured into three blocks. First, general
demographic information is collected, and the task and annotation tool are explained 
to the participant, which involved giving
a brief tutorial and a small guided annotation task, without any assistance features
enabled. This task is only done for training purposes and is not evaluated later. 
Next, the participant has to complete four annotation scenarios, which we 
describe later in this section. After each scenario a short questionnaire is 
conducted, aimed at collecting user opinion, sentiment, and feedback concerning
the scenario they just completed. The last block involves a questionnaire to gather
general feedback and data regarding overall user preferences.

\mypar{Measures}
We measure the effectiveness of assistance features using several metrics aimed at 
\textit{objective} and \textit{subjective} values. For objective values we measure
the time a user takes to annotate a document and the quality of mention, entity, and 
relation annotations, each measured with the $F_1$ score. Subjective values are 
derived from the NASA Task Load 
Index (TLX), which is widely used for measuring the workload during or right after 
performing a task~\cite{hart2006nasa}. The NASA-TLX can be used in many different 
contexts, and was also already used to evaluate information systems
\cite{bagozi2019personalised}. It defines a total of six dimensions 
that measure different aspects of workload. We used the four relevant 
to our study.

\begin{enumerate}[topsep=0pt]
    \item[] \emph{mental demand}: How much the annotator has to focus 
    on the task

    \item[] \emph{uncertainty}: How uncertain the annotator is of their annotations

    \item[] \emph{effort}: How much work is needed to complete the task

    \item[] \emph{frustration}: How frustrated the annotator is with the task
\end{enumerate}
\noindent
We excluded physical and temporal demands, to focus on the subjective 
metrics most relevant to our research 
questions. While \textit{physical} demand is not completely irrelevant (think of mouse 
movements), it is far less informative than the other measures. 
Regarding \textit{temporal} demands we refer to our objective measure of task completion 
time. Note that compared to the original definition of the NASA-TLX, we rephrase 
\textit{performance} to measure the \textit{uncertainty} of an annotator with their 
annotation results. Additionally to the NASA-TLX, we also asked users to share their 
experiences with the tool and assistance features in a questionnaire using 5-point 
Likert items~\cite{joshi2015likert}.

\mypar{Annotation scenarios}
We assess the efficiency and effectiveness of annotators in four scenarios.
Scenario \emph{(A)} \label{scenario:no-assistance} entails no assistance at 
all, besides the workflow defined in Sect.~\ref{sec:workflow} and serves as a 
baseline. Scenario \emph{(B)}\label{scenario:visualization} provides the user 
with visualization of the current annotation, and scenario 
\emph{(C)}\label{scenario:recommendations} gives recommendations for annotations 
made by an artificial intelligence system. Finally, scenario 
\emph{(D)}\label{scenario:combination} combines both assistance features.

Documents in PET contain 168 words on average, which took experts in a preliminary 
experiment as much as 25 minutes to annotate. We therefore decided to instead only 
use fragments of documents, containing two sentences. These fragments were carefully 
selected by measuring the number of mentions, relations, as well as their types. 
We selected fragments from documents \emph{doc-1.2}, \emph{doc-3.6}, \emph{doc-8.3}, 
and \emph{doc-9.2}. Document fragments are part of the supplementary material for 
this paper and available in the repositories mentioned in Sect.~\ref{sec:introduction}.

To avoid 
carry-over effects, i.e., confounding variables such as familiarity with the task after
completing a scenario and therefore performing better in the next one, we use the Balanced 
Latin Square method~\cite{kim2009spreadsheet}. This method systematically produces 
sequences of the scenarios described above, so that each scenario appears as the first 
one in the sequence equally often, as well as two scenarios preceding or succeeding one 
another equally often. Users are assigned a sequence of scenarios in a round-robin fashion, 
compare Fig.~\ref{fig:latin-squares}. 
This setup minimizes the number of scenarios each user has to perform while addressing 
carry-over effects between scenarios, such as increasing familiarization with the 
annotation task.

\section{Results}\label{sec:results}

\begin{figure}[bt]
\centering

\subfloat[]{
    \includegraphics[width=.21\linewidth,valign=c]{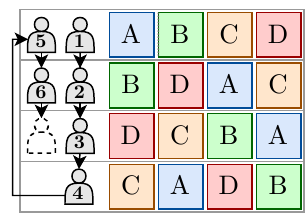}%
    \vphantom{\includegraphics[width=.7\linewidth,valign=c]{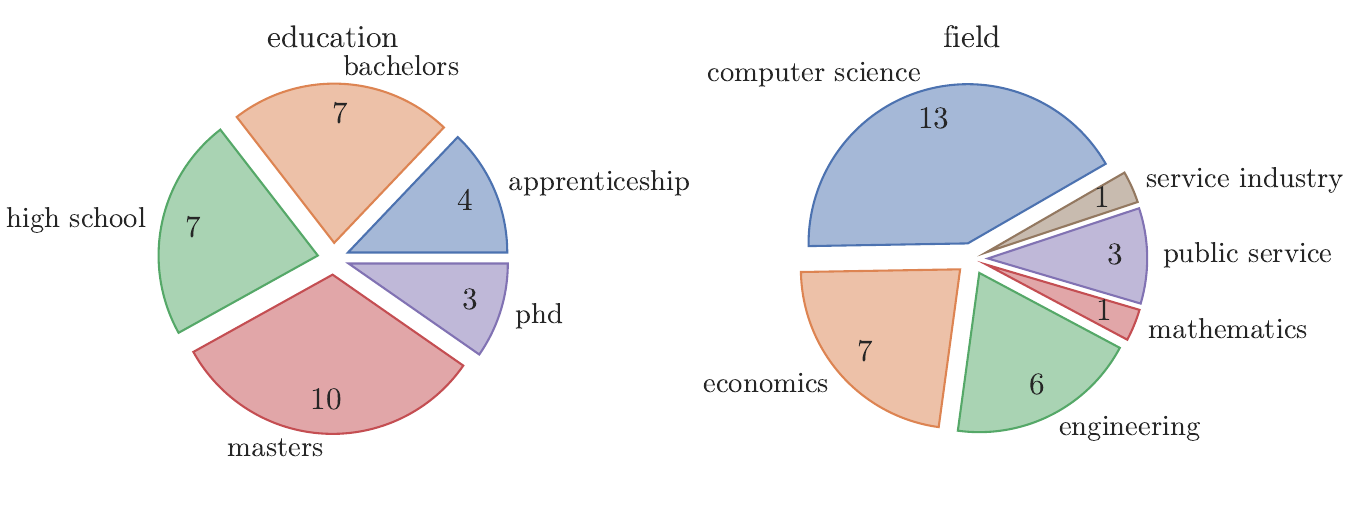}}
    \label{fig:latin-squares}
}%
\hfill%
\subfloat[]{
    \includegraphics[width=.75\linewidth,valign=c]{figures/demographics/combined.pdf}
    \label{fig:demographics-combined}
}

\caption{Assigning annotators to a sequence of scenarios based on a balanced 
Latin square (left), and demographic information about user study participants
(right).}
\end{figure}

In this section we will describe our observations during the experiments 
described in Sect.~\ref{sec:study-design}, starting with an overview of
study participants. We had respondents of various
age, education, and field of work. 39\% have not obtained a 
university degree, or did not pursue higher education, while 
39\% completed either Masters or PhD studies. The majority
(71\%) of participants work in a technical field, i.e., 
computer science, engineering, or mathematics. Fig.~\ref{fig:demographics-combined}
shows a detailed break-down of demographic characteristics of participants. 

\subsection{Subjective Measures}\label{sec:subjective-measures}

As described in Sect.~\ref{sec:study-design}, we measure four subjective
sub-metrics of the NASA-TLX --- mental demand, uncertainty, effort, and frustration 
--- across four different assistance scenarios. We then used a repeated measure 
ANOVA~\cite{bergh1995problems} to find if there are statistically significant 
differences in the four assistance scenarios defined in Sec.~\ref{sec:study-design}. 
A repeated measure ANOVA can be used to test if two ore more non-independent 
samples (measurements) are from the same distribution, measured by 
$p\in{[0,1]}$. In our case, we test for differences in workload between
annotation assistance scenarios. This is the case when we have to reject 
the Null hypothesis, i.e., that the 
measurements have the same variance and mean, which we do if $p<0.05$.

Repeated measures ANOVA assumes sphericity in data, i.e., the difference in 
metrics for all combinations of two scenarios have the same variance. 
This assumption can be tested with Mauchly's test for sphericity
\cite{mauchly1940significance}. Data for three out of four metrics violated 
the assumption of sphericity ($p<0.05$). We use the Greenhouse-Geisser 
correction~\cite{greenhouse1959methods} to account for this. Even then, our 
observations show that each one of the four workload metrics are affected by 
changing how annotators are assisted by our annotation tool and the differences 
are statistically significant with $p<0.001$. 

Since the repeated measures ANOVA indicated a difference in the NASA-TLX metrics
when using different assistance features, we ran six post-hoc tests, looking for 
the differences between each combination of two features, e.g., measurements for 
non-assisted annotation (scenario~\hyperref[scenario:no-assistance]{A} in contrast 
to only recommendations (scenario~\hyperref[scenario:recommendations]{C}). We 
corrected all $p$ values using Bonferroni's method~\cite{ludbrook1998multiple} for 
running multiple tests. Intuitively, running many tests increases the likelihood 
of finding statistically significant differences in one of them, even though there 
is none. This correction multiplies the P-value with the number of tests, to account 
for this increased likelihood. Tab.~\ref{tab:subjective-post-hoc-tests} in 
Sect.~\ref{sec:appendix} reports details.

In summary, no assistance feature at all (scenario~\hyperref[scenario:no-assistance]{A})
is statistically significantly worse than either only recommendations
(scenario~\hyperref[scenario:recommendations]{C}) or both assistance features
combined~\hyperref[scenario:combination]{D}). Surprisingly, assisting annotators
with a visualization of the information they found in the text (i.e., the generated 
graphical process model, scenario~\hyperref[scenario:visualization]{B}) was not found
to help with reducing the workload. 

Compared to no assistance, assisting the annotator with recommendations 
reduced mental demand by $24.7$ ($-34.6\%$), effort by $22.4$ ($-34.2\%$), 
and frustration by $20.5$ ($-51.0\%$). Uncertainty is best lowered by 
combining recommendations with visualizations, which reduces it by $24.4$
($-44.8\%$), according to our observations. Note, that we found no statistically
significant effect on any sub-metric when comparing recommendations (scenario
\hyperref[scenario:recommendations]{C}) to the combination of recommendations and
visualizations (scenario~\hyperref[scenario:combination]{D}). Similarly, we 
could not observe a difference between non-assisted annotation (scenario 
\hyperref[scenario:no-assistance]{A}) and just visualization of annotated information
(scenario~\hyperref[scenario:visualization]{B}). This indicates that only visualizing 
the currently annotated process-relevant information is not enough to 
reduce the workload of the annotation task. Contrary, recommendations 
are a way to reduce it by up to to nearly $50\%$. 
Some limitations apply to our findings concerning the
quality of the graphical process representation, which we discuss in detail
in Sect.~\ref{sec:limitations}. 

\begin{figure}[bt]
    \centering
    \includegraphics[width=\linewidth]{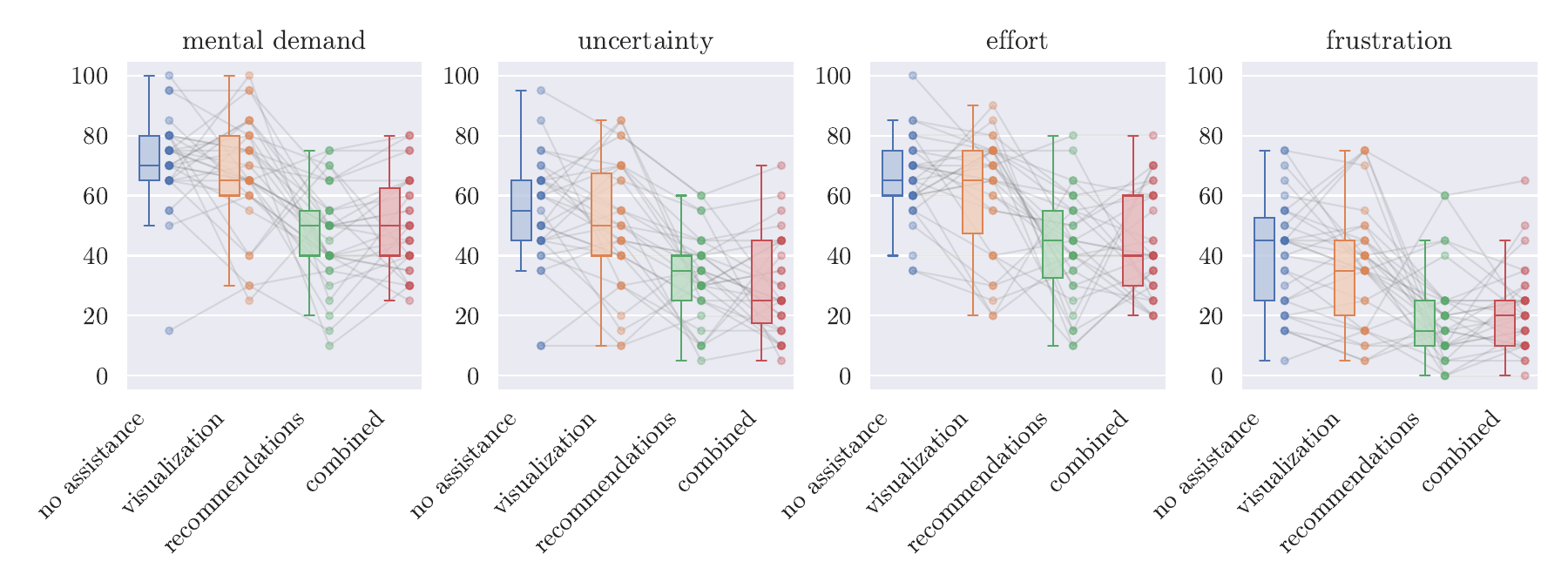}
    \caption{Subjective measures for each of the four scenarios from 
    Sect.~\ref{sec:study-design}.}
    \label{fig:tlx-combined}
\end{figure}

Fig.~\ref{fig:tlx-combined} aggregates our
data for each sub-metric into plots, showing values for each 
participant and scenario as strip plots, where the values of a given participant 
are connected by lines. Additionally the data points are aggregated into 
box plots, showing the data mean, $25^{th}$ and $75^{th}$ percentiles as box,
and the rest of the distribution as whiskers, excluding outliers. The plots
mirror the general observations we drew from Tab.~\ref{tab:subjective-post-hoc-tests},
and shows that recommendations and the combination of both assistance features
help best with reducing the workload of data annotators. 
Overall, the ordering of assistance features in terms of reducing the workload
is obvious from the plots. 
No assistance (scenario~\hyperref[scenario:no-assistance]{A}) 
and visualizations only (scenario~\hyperref[scenario:no-assistance]{B}) share
the spot for least useful, while recommendations and the combination of 
features seem to be equally useful in lowering the workload of 
process information annotation, thus answering research question
\ref{rq:workload-effects}.

\subsection{Objective Measures}\label{sec:objective-measures}

\begin{figure}[bt]
    \centering
    \includegraphics[width=\linewidth]{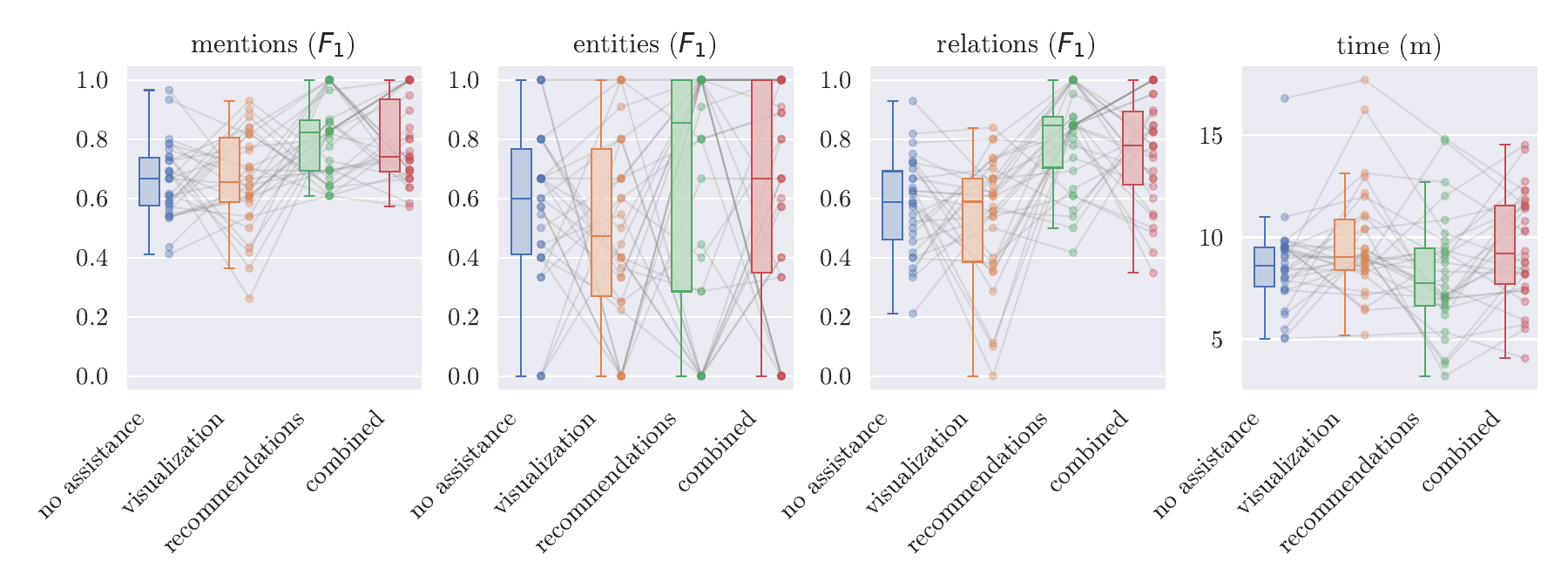}
    \label{fig:objective-metrics-combined}
    \caption{Objective measures for each of the four scenarios from 
    Sect.~\ref{sec:study-design}.}
\end{figure}

As discussed in Sect.~\ref{sec:introduction}, our goal with assisting
data annotators is twofold. The previous section
\ref{sec:subjective-measures} discussed metrics that are subjective,
i.e., are based on the experiences of a data annotator. On the
other hand, assisting annotators also affects the quality 
of annotations. We measured a total of four objective metrics,
which we presented in detail in Sect.~\ref{sec:study-design}.
These are the $F_1$ scores for annotated mentions, entities, and
relations, as well as the total time a given annotator needed to
complete annotating a document fragment. An aggregate of the 
data we obtained is shown in Fig.~\ref{fig:objective-metrics-combined}
as a plot, similar to the one we showed and explained
in Sect.~\ref{sec:subjective-measures}. Detailed results are
listed in Tab.~\ref{tab:objective-post-hoc-tests} in Appendix
\ref{sec:appendix}.

Again, using a repeated measure ANOVA we found significant 
effects on the annotation quality measured in $F_1$ when using 
different assistance features during the annotation of mentions 
($p<0.001$) and relations ($p<0.001$). The annotation quality of 
entities was not affected ($p=0.450$), which may be caused by
the low number of entities\footnote{On average, fragments used in the 
user study only contained one entity that needed resolution, i.e., 
there are at least two entity mentions referring to the same entity.}, 
as well as the fact that we count an entity only as correct, if contains 
all expected mentions. This 
means errors by the annotator during MD propagate to the ER task. 
Furthermore, we did observe a statistically significant difference 
in the time an annotator needs to annotate a document ($p=0.011$), 
but during post hoc tests we could only explain this with a
statistically significant difference between the assistance 
features \textit{visualization} (\hyperref[scenario:recommendations]{B})
and \textit{recommendations} (scenario \hyperref[scenario:recommendations]{C}), 
where the latter speeds up completion times by about 1.5 minutes (see Tab.
\ref{tab:time-post-hoc-tests} in the~\hyperref[sec:appendix]{Appendix}). 

Several participants remarked during the user study, that 
identifying relations and classifying them correctly is a 
very challenging task. These participants were mostly 
inexperienced with BPM and BPMN and thus greatly benefited
from the recommendation assistance feature. We can observe 
this across all participants, as the post hoc tests for
relation annotation quality show. Comparing scenario 
\hyperref[scenario:no-assistance]{A} (no assistance) against 
scenario \hyperref[scenario:recommendations]{C} (recommendations
only), we see a statistically significant ($p=0.001$) 
increase in $F_1$ of $0.224$ ($+38.9\%$). Using the 
visualization does not seem to have a significant
effect compared to no assistance at all ($p=1.000$),
but is significantly worse than using recommendations
($-0.259$, $p<0.001$) or using both assistance features
($-0.204$, $p<0.001$). 

The same analysis can be made for the task of mention
detection. Participants of our user study seem to benefit
most from recommendations, when compared to no assistance
at all ($p<0.001$), with an improvement of $0.141$ 
($+21.4\%$). Visualization
has no statistically significant effect ($p=1.000$) compared
to no assistance at all. Using only visualization has an 
adverse effect compared to just recommendations ($p<0.001$)
with a decrease in $F_1$ of $0.141$. Similar
to Sect.~\ref{sec:subjective-measures}, this effect
can be attributed to limitations in our graphical process
model, which we discuss in Sect.~\ref{sec:limitations}, 
or a user's familiarity with BPMN (Sect.~\ref{sec:experience-results}).

This also answers research question~\ref{rq:quality-effects},
as recommendations seem to be the best choice for improving
the quality of annotations.
Notably, for all three tasks the annotation recommendations 
themselves are of lower quality than the average annotations 
by a human annotator assisted by recommendations (scenario
\hyperref[scenario:recommendations]{C}). Human review
improved the $F_1$ score of annotations by $+0.100$ for MD,
$+0.181$ for ER, and $+0.151\%$ for RE, showing how humans 
assisted by AI-based systems can perform better than each 
part in isolation.

\subsection{Effects of Annotator Experience}\label{sec:experience-results}

We asked participants for their experience with BPMN, measured 
in years. With this information, we now investigate if a user's
experience with BPMN influences how much they can benefit 
from assistance. To this end we split the data into two groups 
--- \textit{experts}, which we define for the purposes of this analysis 
as participants with at least one year 
of BPMN modeling experience, and \textit{novices}, which are the 
remaining study participants.
This split results in 10 experts and 18 novices. 
We hypothesize that annotations by \textit{novices} are worse in 
terms of $F_1$ score compared to those by \textit{experts}. 
Tab.~\ref{tab:independent-samples-t-test} lists results 
for an independent samples T-Test.

\begin{table}[bt]
    \sisetup{table-align-text-post=false, table-space-text-post={\,\%}}
    
    \centering
    \caption{Independent Samples T-Test for the hypothesis that objective scores for annotations 
    by \textit{novices} are lower than those by \textit{experts}.}
	\label{tab:independent-samples-t-test}
 
    \begin{tabularx}{0.75\textwidth}{rLS[table-format=4.3]S[table-format=4.3]S[table-format=3.3]l}

    \toprule

                                    &                   & {\thead[c]{t}}    & {\thead[c]{df}}   & \multicolumn{2}{c}{\thead[r]{${p}^{a}$}} \\

    \toprule
     
    \rowheader{4}{mentions}         & no assistance     & -1.950            & 20.691            & 0.032             & $^{*}$   \\
                                    & recommendations   &  0.023            & 23.041            & 0.509             &          \\
                                    & visualization     & -1.166            & 23.984            & 0.128             &          \\
                                    & combined          & -1.463            & 24.491            & 0.078             &          \\

    \cmidrule{1-6}
    
    \rowheader{4}{relations}        & no assistance     & -1.800            & 25.072            & 0.042             & $^{*}$   \\
                                    & recommendations   &  0.664            & 21.132            & 0.743             &          \\
                                    & visualization     & -1.182            & 28.000            & 0.124             &          \\
                                    & combined          & -0.590            & 20.711            & 0.281             &          \\

    \bottomrule

    \noalign{\vskip 1mm}
    
    \multicolumn{6}{l}{$^{*}$, $^{**}$, $^{***}$ statistically significant results of increasing degrees.} \\
    \multicolumn{6}{l}{$^{a}$P-value following Welch's test.}
    
    \end{tabularx}
\end{table}

We can confirm our assumption that BPMN experience improves the quality 
of mention (MD) and relation annotations (RE), for un-assisted 
annotation (scenario~\hyperref[scenario:no-assistance]{A}).
In all assisted scenarios (\hyperref[scenario:recommendations]{B},
\hyperref[scenario:recommendations]{C}, \hyperref[scenario:recommendations]{D})
we have to reject our hypothesis, i.e., \textit{novices} no 
longer produce worse annotations than \textit{experts}, from which 
we infer that the two assistance features can indeed bridge the gap 
in annotation quality caused by differences in experience with BPMN. 
We therefore answer research question~\ref{rq:bridge-gap} with 
\textit{annotation recommendations}, \textit{visualizations}, and
\textit{combined assistance}.

\section{Conclusion}\label{sec:conclusion}

In this section we will reiterate the core contribution of this paper, 
the limitations of the user study we conducted, and we describe our plans
for future work.

\mypar{Core contributions}\label{sec:contributions}
This paper presents an in-depth exploration on the usefulness of 
two features for assisting data annotators in the domain of business
process information extraction. A user study with 31
participants shows that annotation recommendations reduce certain
workload aspects by up to $-51.0\%$ (\ref{rq:workload-effects}).
We find that recommendations obtained by a system based on machine
learning improve annotation quality as much as $+38.9\%$ (\ref{rq:quality-effects}). 
The same recommendations bridge the gap in annotation quality between beginner 
and expert annotators, promising easier assembly of annotation teams
by means of shorter training times (\ref{rq:bridge-gap}).
We make all data and code publicly available.

\mypar{Limitations}\label{sec:limitations}
This work is still limited in several ways. First, we focused our 
study on two assistance features, to ensure its feasibility, while 
also guaranteeing methodological correctness. 
Investigating more assistance features would either
increase participation times, or limit each participation to 
a subset of scenarios. 

Next, while we could not observe statistically significant effects
of any assistance features on the quality of entity resolution 
annotations, we cannot eliminate the possibility that this caused by errors 
propagated from the MD task. 

Our automated method used for generating and layouting a graphical
process model from the process information (annotations) used for the
visualization assistance feature has limitations in terms of structure, 
accuracy stemming from the employed heuristics, and clarity of generated
labels. This may affect its usefulness, as these limitations may make
the graphical model harder to understand, especially for untrained
annotators.

Finally, we only present and analyze a sub-set of the data collected 
during the user study. For example, we recorded all user interaction 
with the tool, such as when a recommended annotation is discarded, 
or a new annotation is created. These logs constitute valuable data
for improving the workflow for annotating process relevant data
in textual process descriptions.

\mypar{Future work}\label{sec:future-work}
Our future work is mainly concerned with eliminating the limitations 
we discussed in the previous section (\ref{sec:limitations}). As such
we plan to improve the implementation of our annotation tool, e.g., 
improve the way relations are displayed. We also plan to extend our 
analysis of the data we already obtained during this user study, e.g., 
by evaluating the interaction logs. This data can be very valuable
to learn how annotators interact with the annotation tool, and give
indications on how to improve the workflow, or which parts of the
interface are still unintuitive.

Furthermore, we want to explore better annotation recommendation 
methods, as this feature seems to have a consistently positive effect.
We plan to evaluate integrating incremental training, as soon as
annotators have submitted a document. 

Finally we would like to extend the user study to new assistance
features, in addition to comparing different workflows and user 
interface options. The initial findings regarding how 
experts benefit in different ways from assistance features, 
compared to novice users, motivate us to conduct a targeted study
to find ways to properly assist users of different experience
levels.

\begin{appendix}
    \section*{Appendix}
\label{sec:appendix}

\begin{table}[H]
    \sisetup{table-align-text-post=false, table-space-text-post={\,\%}}
    
    \centering
    \caption{Post hoc comparisons of assistance features on objective 
    metrics. Largest statistically significant absolute difference to 
    unassisted annotation is set in \textbf{bold}.
    We abbreviate mean difference with MD and standard error with SE.}
    \label{tab:objective-post-hoc-tests}
    \begin{tabularx}{\textwidth}{cLLS[table-format=3.3]S[table-format=3.3]S[table-format=3.4]rl}

    \toprule

                                    &                   &                   & {\thead[c]{MD}}  & {\thead[c]{SE}}   & {\thead[c]{t}}   & \multicolumn{2}{c}{\thead[r]{${p_{bonf}}^{a}$}} \\

    \toprule
     
    \rowheader{7}{mentions}         & no assistance     & recommendations   & \textbf{-0.141}               & 0.037             & -3.774            &                                 $0.002$               & $^{**}$   \\
                                    &                   & visualization     & -0.000                        & 0.037             & -0.008            & $1.000$               &           \\
                                    &                   & both              & -0.132                        & 0.037             & -3.529            & $0.014$               & $^{**}$   \\
    \cmidrule{2-8}
                                    & recommendations   & visualization     & 0.141                         & 0.037             & 3.766             & $0.005$               & $^{**}$   \\
                                    &                   & both              & 0.009                         & 0.037             & 0.244             & $1.000$               &           \\
    \cmidrule{2-8}
                                    & visualization     & both              & -0.132                        & 0.037             & -3.522            & $0.004$               & $^{**}$   \\

    \cmidrule{1-8}
    
    \rowheader{7}{entities}         & no assistance     & recommendations   & -0.068                        & 0.097             & -0.706            &                                 $1.000$               & \\
                                    &                   & visualization     & 0.080                         & 0.097             & 0.825             & $1.000$               & \\
                                    &                   & both              & -0.038                        & 0.097             & -0.398            & $1.000$               & \\
    \cmidrule{2-8}
                                    & recommendations   & visualization     & 0.148                         & 0.097             & 1.532             & $0.775$               & \\
                                    &                   & both              & 0.030                         & 0.097             & 0.309             & $1.000$               & \\
    \cmidrule{2-8}
                                    & visualization     & both              & -0.118                        & 0.097             & -1.223            & $1.000$               & \\

    \cmidrule{1-8}
    
    \rowheader{7}{relations}        & no assistance     & recommendations   & \textbf{-0.224}               & 0.049             & -4.524            &                                 $<.001$               & $^{***}$  \\
                                    &                   & visualization     & 0.025                         & 0.049             & 0.842             & $1.000$               &           \\
                                    &                   & both              & -0.179                        & 0.049             & -3.691            & $0.002$               & $^{**}$  \\
    \cmidrule{2-8}
                                    & recommendations   & visualization     & 0.259                         & 0.049             & 5.367             & $<.001$               & $^{***}$  \\
                                    &                   & both              & 0.055                         & 0.049             & 0.834             & $1.000$               &           \\
    \cmidrule{2-8}
                                    & visualization     & both              & -0.204                        & 0.049             & -4.533            & $<.001$               & $^{***}$  \\

    \bottomrule
    
    \noalign{\vskip 1mm}

    \multicolumn{8}{l}{$^{*}$, $^{**}$, $^{***}$ statistically significant results of increasing degrees.} \\
    \multicolumn{8}{l}{$^{a}$P-value adjusted for comparing a family of six using Bonferroni correction.}
    
    \end{tabularx}
\end{table}

\begin{table}[H]
    \sisetup{table-align-text-post=false, table-space-text-post={\,\%}}
    
    \centering
    \caption{Post hoc comparisons of assistance features on completion time.
    We abbreviate mean difference with MD and standard error with SE.}
    \label{tab:time-post-hoc-tests}
    \begin{tabularx}{\textwidth}{cLLS[table-format=3.3]S[table-format=3.3]S[table-format=3.4]rl}

    \toprule

                                    &                   &                   & {\thead[c]{MD}}  & {\thead[c]{SE}}   & {\thead[c]{t}}   & \multicolumn{2}{c}{\thead[r]{${p_{bonf}}^{a}$}} \\

    \toprule
    
    \rowheader{7}{time (s)}         & no assistance     & recommendations   & 23.778                        & 30.528            & 0.779             &                                 $1.000$               &   \\
                                    &                   & visualization     & -67.621                       & 30.528            & -2.215            & $0.176$               &   \\
                                    &                   & both              & -56.082                       & 30.528            & -1.837            & $0.418$               &   \\
    \cmidrule{2-8}
                                    & recommendations   & visualization     & -91.399                       & 30.528            & -2.994            & $0.022$               & $^{*}$   \\
                                    &                   & both              & -79.860                       & 30.528            & -2.616            & $0.063$               &   \\
    \cmidrule{2-8}
                                    & visualization     & both              & 11.539                        & 30.528            & 0.378             & $1.000$               &   \\

    \bottomrule
    
    \noalign{\vskip 1mm}

    \multicolumn{8}{l}{$^{*}$, $^{**}$, $^{***}$ statistically significant results of increasing degrees.} \\
    \multicolumn{8}{l}{$^{a}$P-value adjusted for comparing a family of six using Bonferroni correction.}
    
    \end{tabularx}
\end{table}

\begin{table}[H]
    \sisetup{table-align-text-post=false, table-space-text-post={\,\%}}
    
    \centering
    \caption{Post hoc comparisons of assistance features on subjective 
    metrics. Largest statistical significant absolute difference to 
    unassisted annotation for a given metric is set in \textbf{bold}.
    We abbreviate mean difference with MD and standard error with SE.}
    \label{tab:subjective-post-hoc-tests}
    \begin{tabularx}{\textwidth}{cLLS[table-format=3.3]S[table-format=3.3]S[table-format=3.4]rl}

    \toprule

                                    &                   &                   & {\thead[c]{MD}}  & {\thead[c]{SE}}   & {\thead[c]{t}}   & \multicolumn{2}{c}{\thead[r]{${p_{bonf}}^{a}$}} \\

    \toprule
     
    \rowheader{7}{mental\\demand}   & no assistance     & recommendations   & \textbf{24.677}               & 3.521             & 7.009             & $<.001$               & $^{***}$  \\
                                    &                   & visualization     & 4.194                         & 3.521             & 1.191             & $1.000$               &           \\
                                    &                   & both              & 21.290                        & 3.521             & 6.047             & $<.001$               & $^{***}$  \\
    \cmidrule{2-8}
                                    & recommendations   & visualization     & -20.484                       & 3.521             & -5.818            & $<.001$               & $^{***}$  \\
                                    &                   & both              & -3.387                        & 3.521             & -0.962            & $1.000$               &           \\
    \cmidrule{2-8}
                                    & visualization     & both              & 17.097                        & 3.521             & 4.856             & $<.001$               & $^{***}$ \\

    \cmidrule{1-8}
    
    \rowheader{7}{uncertainty}      & no assistance     & recommendations   & 21.129                        & 3.558             & 5.938             & $<.001$               & $^{***}$  \\
                                    &                   & visualization     & 4.677                         & 3.558             & 1.314             & $1.000$               &           \\
                                    &                   & both              & \textbf{24.355}               & 3.558             & 6.844             & $<.001$               & $^{***}$  \\
    \cmidrule{2-8}
                                    & recommendations   & visualization     & -16.452                       & 3.558             & -4.623            & $<0.001$              & $^{***}$  \\
                                    &                   & both              &  3.226                        & 3.558             & 0.907             & $1.000$               &           \\
    \cmidrule{2-8}
                                    & visualization     & both              & 19.677                        & 3.558             & 5.530             & $<.001$               & $^{***}$   \\

    \cmidrule{1-8}
    
    \rowheader{7}{effort}           & no assistance     & recommendations   & \textbf{22.419}               & 3.710             & 6.043             & $<.001$               & $^{***}$  \\
                                    &                   & visualization     & 5.000                         & 3.710             & 1.348             & $1.000$               &           \\
                                    &                   & both              & 21.129                        & 3.710             & 5.695             & $<.001$               & $^{***}$  \\
    \cmidrule{2-8}
                                    & recommendations   & visualization     & -17.419                       & 3.710             & -4.695            & $<.001$               & $^{***}$  \\
                                    &                   & both              & -1.290                        & 3.710             & -0.348            & $1.000$               &           \\
    \cmidrule{2-8}
                                    & visualization     & both              & 16.129                        & 3.710             & 4.347             & $<.001$                & $^{***}$   \\

    \cmidrule{1-8}
    
    \rowheader{7}{frustration}      & no assistance     & recommendations   & \textbf{20.484}               & 3.655             & 5.604             & $<.001$               & $^{***}$  \\
                                    &                   & visualization     & 3.548                         & 3.655             & 0.971             & $1.000$               &           \\
                                    &                   & both              & 18.548                        & 3.655             & 5.074             & $<.001$               & $^{***}$  \\
    \cmidrule{2-8}
                                    & recommendations   & visualization     & -16.935                       & 3.655             & -4.633            & $<.001$               & $^{***}$  \\
                                    &                   & both              & -1.935                        & 3.655             & -0.530            & $1.000$               &           \\
    \cmidrule{2-8}
                                    & visualization     & both              & 15.000                        & 3.655             & 4.104             & $<.001$               & $^{***}$  \\

    \bottomrule
    
    \noalign{\vskip 1mm}

    \multicolumn{8}{l}{$^{*}$, $^{**}$, $^{***}$ statistically significant results of increasing degrees.} \\
    \multicolumn{8}{l}{$^{a}$P-value adjusted for comparing a family of six using Bonferroni correction.}
    
    \end{tabularx}
\end{table}
\end{appendix}

\begin{credits}
    \subsubsection{\discintname}
    The authors have no competing interests to declare that are
    relevant to the content of this article.
\end{credits}

\bibliographystyle{splncs04}
\bibliography{bibliography}

\end{document}